\pdfoutput=1
\documentclass[11pt]{article}

\usepackage[final]{ACL2023}

\usepackage{times}
\usepackage{latexsym}
\usepackage{graphicx}
\usepackage{hyperref}
\usepackage{multirow}
\usepackage{amsmath}

\usepackage[T1]{fontenc}
\usepackage[utf8]{inputenc}

\usepackage{microtype}

\usepackage{inconsolata}

\title{IndoNLP 2025: Shared Task on Real-Time Reverse Transliteration for
Romanized Indo-Aryan languages}

\author{
    Deshan Sumanathilaka$^{1}$$^{*}$, 
    Isuri Anuradha$^{2}$, 
    Ruvan Weerasinghe$^{3}$,
    Nicholas Micallef$^{1}$,
    Julian Hough$^{1}$
     \\
    $^1$Swansea University, Wales, UK \\
    $^2$Lancaster University, UK \\
    $^3$Informatics Institute of Technology, Sri Lanka \\
    *Corresponding email :\text{deshankoshala@gmail.com} 
}

\begin{document}
\maketitle
\begin{abstract}
The paper overviews the shared task on Real-Time Reverse Transliteration for Romanized Indo-Aryan languages. It focuses on the reverse transliteration of low-resourced languages in the Indo-Aryan family to their native scripts. Typing Romanized Indo-Aryan languages using ad-hoc transliterals and achieving accurate native scripts are complex and often inaccurate processes with the current keyboard systems. This task aims to introduce and evaluate a real-time reverse transliterator that converts Romanized Indo-Aryan languages to their native scripts, improving the typing experience for users. Out of 11 registered teams, four teams participated in the final evaluation phase with transliteration models for  Sinhala, Hindi and Malayalam. These proposed solutions not only solve the issue of ad-hoc transliteration but also empower low-resource language usability in the digital arena.
\end{abstract}

\section{Introduction}

Languages transcend mere systems of communication. They are profound reflections of the cultures they represent. Embedded within each language are the accumulated wisdom, traditions, beliefs, and historical narratives cherished by its speakers. South Asia, a region renowned for its cultural richness and linguistic diversity, is home to a vast array of languages that serve not only as tools for communication but also as vibrant expressions of its multifaceted heritage.

Among these, the Indo-Aryan languages hold a significant place, forming part of the larger Indo-Iranian branch of the Indo-European language family. As of the early 21st century, the Indo-Aryan languages (sometimes called Indic languages) have more than 800 million speakers, primarily concentrated east of the Indus River in Bangladesh, North India, Eastern Pakistan, Sri Lanka, Maldives, and Nepal \cite{kj2024decoding}. Additionally, large immigrant and expatriate Indo-Aryan-speaking communities can be found in Northwestern Europe, Western Asia, North America, the Caribbean, Southeast Africa, Polynesia, and Australia. There are over 200 known Indo-Aryan languages, further emphasizing the immense diversity within this linguistic group \cite{talukdar2023parts}. Historically, the Indo-Aryan languages have evolved from Old Indo-Aryan (Sanskrit) through Middle Indo-Aryan (Prakrits) to their modern forms, such as Hindi, Bengali, Gujarati, Punjabi, and others. This historical evolution underscores their linguistic richness and role as carriers of cultural and historical narratives \cite{masica1993indo}.

In contrast, the Dravidian languages constitute a distinct language family, primarily spoken in South India and parts of Eastern and Central India, as well as in northeastern Sri Lanka, Pakistan, Nepal, Bangladesh, and among diaspora communities worldwide. The Indo-Aryan and Dravidian languages illustrate South Asia's profound linguistic and cultural complexity, weaving a vibrant tapestry of communication and heritage that resonates far beyond the region \cite{de2019survey}. With the advancement of digital technologies, communication on social media has become increasingly prominent. Social media users often employ both native scripts and Romanized versions of their native languages for accessible communication and compatibility with digital platforms. However, the transliteration of Indo-Aryan languages into native scripts remains under-explored, mainly due to a scarcity of language resources needed for their development. While numerous rule-based systems have been implemented to transliterate Indo-Aryan languages, significant challenges persist in addressing ad-hoc transliteration effectively. Social media users frequently rely on abbreviated and informal typing styles to communicate in their native languages. This variability in linguistic representation complicates the verification of hate speech and misinformation, as the diverse formats pose substantial challenges for detection and analysis \cite{sumanathilaka_swa-bhasha_2023}.

The key contributions of this shared task are :
\begin{itemize}
    \item Introducing novel models for Romanized scripts to Native Indo-Aryan Languages Transliteration.
    \item Introducing a standard evaluation dataset for five languages: Sinhala, Hindi, Bengali, Gujarati and Malayalam.
\end{itemize}

Moving forward, the paper will discuss related works, shared task overview, results and discussion along with the conclusions.

\section{Related Works}

Early back-transliteration methods for Indo-Aryan languages heavily relied on rule-based approaches, employing predefined character mappings and linguistic rules. For example, \citet{vidanaralage_sinhala_2018} introduced a rule-based method for Sinhala transliteration using transliteration and phoneme rule bases. Similarly, \citet{tennage_transliteration_2018} utilized character mapping tables for converting Sinhala into English. These methods demonstrated effectiveness in controlled environments but often faltered with informal Romanized text, such as Singlish, which includes significant variability and ambiguity. Researchers have combined rule-based methods with statistical models to overcome the limitations of purely rule-based systems. \citet{liwera_combination_2020} applied a trigram model trained on Romanized Sinhala YouTube comments, integrated with rule-based techniques, to address transliteration challenges. Hybrid models have shown promise but struggle with contextual dependencies and ambiguities in informal text. Classical machine learning models were explored to transliterate Hindi and Marathi to English efficiently \cite{rathod2013hindi}.

Recent advancements in deep learning have revolutionized transliteration tasks. Neural network-based sequence-to-sequence models have gained traction, particularly those utilizing recurrent architectures. \citet{kunchukuttan2015brahmi} introduced Brahmi-Net, a transliteration system leveraging statistical approaches for 18 Indo-Aryan languages. \citet{nanayakkara_context_2022} employed a Bidirectional LSTM and LSTM for Sinhala transliteration, incorporating character-level context. Transformer architectures have emerged as state-of-the-art for transliteration due to their ability to handle long-range dependencies with self-attention mechanisms. \citet{zohrabi2023borderless} employed Transformer-based models for Azerbaijani transliteration while \citet{madhani2023aksharantar} introduced IndicXlit, a multilingual system leveraging Transformer architecture. Pretrained multilingual models like mT5 and M2M100 have also been fine-tuned for transliteration tasks, treating transliteration as a translation problem. A deep learning approach proposed by \cite{deselaers2009deep} used a deep belief network for Arabic-English transliteration, and a few other Arabizi to Arabic transliteration are explored \cite{masmoudi2019transliteration,al2014automatic}.

Hybrid approaches combining rule-based, statistical, and deep learning methods have demonstrated increased accuracy and robustness. \citet{athukorala_swa_2022} proposed a hybrid approach for Singlish transliteration, integrating rule-based methods with fuzzy logic and achieving a word-level accuracy of 0.64. This work has been further extended to Ngram with Rule based approach by \citet{sumanathilaka_swa-bhasha_2023} and later \citet{dharmasiri_swa_2024} by using a GRU for efficient transliteration. TAMZHI by \citet{mudiyanselage2024tamழி} developed a back-transliteration system for Romanized Tamil, achieving 93\% character-level accuracy and 70\% word-level accuracy. Pix2Pix Generative Adversarial Network for transliterating ancient Indian scripts, demonstrating the adaptability of GANs for image-to-text transliteration were further explored \cite{sharma_ancient_2023}.

The availability of datasets like Dakshina \cite{roark2020processing} and Aksharantar\cite{madhani2023aksharantar}  has significantly enhanced transliteration model training.  \citet{sumanathilaka_swa-bhasha_2024} has built a large corpus of data which consists of Sinhala and its Romanized Sinhala patterns along with a dataset to handle Word sense disambiguation in Romanized Sinhala scripts. These large-scale resources provide robust transliteration pairs, enabling models to learn from diverse and complex Romanized text patterns. However, the lack of standardized rules for transliteration leads to ad-hoc spelling variations, including the omission or alteration of vowels, inconsistent phonetic mappings, and user-specific adaptations. These inconsistencies create significant challenges in accurately converting Romanized text into native scripts. Existing keyboard systems and transliteration tools frequently fall short of addressing these complexities, often producing inaccurate or unintelligible outputs \cite{mammadzada2023review}. This results in frustration for users, reduces the accessibility of native scripts, and hinders seamless communication in their preferred languages. Therefore, the demand for a robust and reliable solution to this problem is practical and urgent.

\section{Shared Task Overview}

\subsection{Shared Task Definition}
The widespread use of Romanized Indo-Aryan languages and native languages expressed using the English alphabet has become a prevalent mode of communication in digital spaces. This practice is especially common in informal settings such as social media and messaging platforms \cite{yadav2023different}. However, the absence of standardized transliteration rules results in inconsistent spellings, phonetic mismatches, and user-specific adaptations, creating significant challenges for accurately converting Romanized text to native scripts and highlighting the need for reliable solutions.

To address these challenges, this shared task introduces a real-time reverse transliterator capable of converting Romanized Indo-Aryan language input into its accurate native script equivalent. The solution aims to improve the transliteration process significantly, providing a smoother and more user-friendly typing experience for speakers of Indo-Aryan languages who rely on Romanized input. By enabling accurate and efficient native script outputs, the project seeks to bridge the gap between linguistic preferences and digital accessibility.

Participants in this shared task are required to develop models capable of:
\begin{itemize}
    \item Accepting Romanized text input for one or more Indo-Aryan languages.
    \item Producing accurate native script output in real-time or near-real-time settings.
    \item Handling ad-hoc transliterations, including inputs with inconsistent or missing vowels, abbreviations, and phonetic variations.
    \item Operating efficiently under constraints such as limited computational resources.
\end{itemize}

\begin{figure}[h!]
    \centering
    \includegraphics[width=0.5\textwidth]{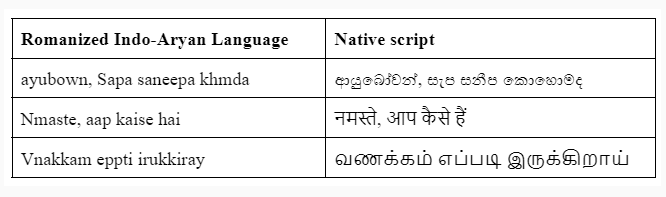} 
    \caption{Example cases of Input and Output.}
    \label{fig:examples}
\end{figure}

\subsection{Objectives of shared Task}

The primary objective of this project is to develop a system that accurately converts Romanized Indo-Aryan language text into the native script in real time, ensuring high precision and usability. The system is designed to handle transliterations with or without vowels, effectively resolving ambiguities arising from such variations. Handling such ambiguities has been a major drawback in the current transliteration system used in commercial platforms \cite{abeysiriwardana2024survey}. By addressing the limitations and inaccuracies inherent in existing keyboard systems, the project aims to provide users with a seamless and efficient typing experience, enabling reliable communication in native scripts while accommodating the diverse and ad-hoc transliteration styles commonly used in digital interactions. Detailed input and output are shown in the Figure\ref{fig:examples}.

\subsection{Key Challenges to Address}

Transliterating Romanized Indo-Aryan languages into native scripts presents several challenges that stem from linguistic and technological complexities. Ad-hoc transliterations are common, with users employing inconsistent spellings, including omitting or substituting vowels, which complicates accurate mapping to native scripts. The lack of standardization in Romanized spellings, which are often user-specific and devoid of formal rules, further contributes to significant variability. Phonetic ambiguity adds another layer of difficulty, as multiple Romanized representations can correspond to a single native script word and vice versa. Additionally, resource scarcity poses a major hurdle, with Indo-Aryan languages lacking comprehensive datasets for training and evaluating effective transliteration systems. To ensure usability, addressing these challenges while meeting real-time constraints and delivering low-latency, high-accuracy transliteration for seamless user interaction is essential.

\subsection{Datasets}
\subsubsection{Training datasets}
The proposed study leverages multiple comprehensive datasets for transliteration tasks. The first dataset, the \textbf{Dakshina Dataset}\footnote{\url{https://github.com/google-research-datasets/dakshina}}, provides a rich collection of text in native scripts and their Romanized counterparts for various Indo-Aryan languages. This dataset is particularly useful for training and evaluating models on diverse transliteration patterns. This dataset was mostly used by the participants to train and test their models. The second dataset, the \textbf{Aksharantar Dataset}\footnote{\url{https://github.com/AI4Bharat/IndicXlit}}, offers an extensive repository of Romanized and native script pairs, focusing on Indian languages. 

Additionally, for Sinhala, the \textbf{Swa-Bhasha Dataset}\footnote{\url{https://github.com/Sumanathilaka/Swa-Bhasha-Sinhala-Singlish-Dataset}} is proposed, which contains a wide range of transliteration scripts specifically tailored for Romanized Sinhala to Sinhala. This dataset comprises four distinct types: unique words, Sinhala - Romanized Sinhala ad-hoc transliterations, Romanized Sinhala - Sinhala social media datasets, and the WSD Romanized Sinhala - Sinhala Transliteration Dataset. These datasets provide a robust foundation for developing and benchmarking the transliteration system, addressing challenges such as ad-hoc spellings, phonetic ambiguities, and resource scarcity. This dataset has been used by Team Vectora for their training process.

\subsubsection{Test datasets}

This test dataset has been created and augmented specifically for the \textbf{IndoNLP Shared Task}\footnote{\url{https://github.com/IndoNLP-Workshop/IndoNLP-2025-Shared-Task/}}. Please note that some data records are a combination of existing datasets that are publicly available for the respective languages. The augmentation process involved generating new data samples based on these existing resources while ensuring data diversity and relevance to the task. The results are presented using Word Error Rate (WER), Character Error Rate (CER), and BiLingual Evaluation Understudy (BLEU), which are standard metrics to evaluate the quality of transliteration systems, measuring errors and assessing linguistic fidelity. The data distribution is presented in the Table \ref{tab:typing-patterns}.

\begin{table*}[h!]
\centering
\begin{tabular}{|c|c|c|}
\hline
\textbf{Language} & \textbf{Test Set 1 : General Typing Patterns} & \textbf{Test Set 2 : Ad-hoc Typing Patterns} \\ \hline
Sinhala           & 10,000                                       & 5,000                                      \\ \hline
Bengali           & 10,000                                       & 5,000                                      \\ \hline
Gujarati          & 5,000                                        & 5,000                                      \\ \hline
Hindi             & 5,000                                        & 5,000                                      \\ \hline
Malayalam         & 10,000                                       & 5,000                                      \\ \hline
\end{tabular}
\caption{Test sets for general and ad-hoc typing patterns across languages.}
\label{tab:typing-patterns}
\end{table*}

\subsection{Participant's Systems}

\begin{table*}[h]
\centering
\begin{tabular}{|l|l|l|}
\hline
\textbf{Team Name} & \textbf{Affiliation} & \textbf{Language} \\
\hline
IndiDataMiner & Indian Institute of Technology Guwahati, India & Hindi \\
\hline
Vectora & Informatics Institute of Technology, Sri Lanka & Sinhala \\
\hline
MoraCSE & University of Moratuwa, Sri Lanka & Sinhala \\
\hline
NexText & Digital University Kerala, India & Malayalam \\
\hline
\end{tabular}
\caption{Team Information and Language Distribution}
\label{tab:team_info}
\end{table*}

Table \ref{tab:team_info} presents the registered team information, their affiliations, and the primary languages they worked on. The discussion below provides an analysis of the users' proposed solutions.

The \textbf{IndiDataMiner} team \cite{kumar2025team} at the Indian Institute of Technology Guwahati proposed a sentence-level back-transliteration approach using the LLaMa 3.1 model for Hindi. Their approach addresses the challenges posed by the increasing use of Romanized typing for Indo-Aryan languages on social media, which often lacks standardization and results in loss of linguistic richness. To resolve the ambiguities in Romanized Hindi text, they leveraged fine-tuning with the Dakshina dataset. The team's approach includes both zero-shot learning and fine-tuning techniques to enhance the model's transliteration capabilities. The model was trained using the Dakshina dataset, which provides a parallel corpus of 12 Indian languages, including Hindi. The dataset was formatted to fit the Alpaca prompt structure, with the instruction to transliterate Romanized Hindi back to Devanagari script, ensuring consistency across training and testing. They used the LLaMA 3.1 8B model, a large-scale transformer-based architecture, which was optimized for causal language modeling and enhanced with Low-Rank Adaptation (LoRA) and 4-bit quantization. The training process utilized the SFTTrainer class from the trl library and the Unsloth framework for 4-bit quantization, which improved memory efficiency. The fine-tuned model demonstrated significant improvements in transliteration accuracy, achieving high BLEU scores on the Hindi test dataset.

\textbf{Team Vectora}'s \cite{perera2025indonlp} approach to Sinhala back-transliteration is a context-aware system that combines multiple techniques to handle the complexities of "Singlish". Their method employs several sophisticated components that work together to achieve accurate transliteration. At the core of their system is a dictionary-based mapping approach that uses an ad-hoc transliteration dictionary to map common Singlish words to their Sinhala equivalents. This dictionary is particularly effective at handling frequent words and their different possible Sinhala representations. To accommodate the complexity of the language, the dictionary incorporates multiple mappings for ambiguous words. For handling words not found in the dictionary (out-of-vocabulary words), the system implements rule-based transliteration based on Sinhala phonetic patterns. This component proves essential for processing words that may not be common or are absent from their dictionary. To resolve lexical ambiguities where one Romanized word can map to multiple Sinhala words, the system employs a BERT-based language model to analyze the sentence-level context and select the most appropriate Sinhala word.
The contextual disambiguation process works through a sophisticated sequence of operations. The system first generates all possible sentences by filling the masked ambiguous words with candidate Sinhala words. These generated sentences are then evaluated using a BERT model configured for Masked Language Modeling (MLM). The final transliterated output is determined by selecting the sentence that achieves the highest probability score. To address processing time challenges, the system incorporates several optimization strategies. These include reducing candidate words for ambiguous words through a filtering mechanism and implementing sentence chunking based on the number of BERT calls. The effectiveness of the entire system is rigorously evaluated using multiple metrics.

The \textbf{Team MoreCSE} \cite{de2024sinhala} explored two distinct approaches for Sinhala transliteration: a rule-based system and a deep learning-based system. The rule-based approach implements a systematic method using predefined linguistic rules to map Latin script (Singlish) to Sinhala script. This system operates through a character-by-character matching strategy, processing each input word by matching the longest possible substring (up to three characters) with rules in a transliteration table. The system follows a straightforward logic: when a match is found, the corresponding Sinhala character is appended to the result; if no match is found, the character is added unchanged. This method builds upon and enhances a previous rule-based system by incorporating additional rules for two and three-character mappings. Their deep learning approach takes a fundamentally different path by modelling transliteration as a translation task, leveraging a Transformer-based encoder-decoder architecture. The team fine-tuned the M2M100 model, a pre-trained multilingual sequence-to-sequence model, by treating Romanized Sinhala as the English input and the Sinhala script as the target language. This implementation utilizes the M2M100's tokenizer and learns the intricate relationships between the two language pairs. The deep learning method offers several key advantages: it enables context-based generation, eliminates the need for manual rule definition, and can effectively handle code-mixed and code-switched cases through expanded training data. The model's training was conducted on a substantial dataset comprising 10k parallel data points. Their analysis revealed that while the deep learning approach demonstrated superior robustness in handling language variability, it proved less computationally efficient than its rule-based counterpart.

\begin{table*}[h]
\centering
\begin{tabular}{|l|l|c|c|c|c|}
\hline
\textbf{Team} & \textbf{Model} & \textbf{Test} & \textbf{WER} & \textbf{CER} & \textbf{BLEU} \\
\hline
\multirow{4}{*}{Team Vectora} & \multirow{2}{*}{Sinhala BERT} & Test 1 & 0.0886 & 0.0200 & 0.9115 \\
 & & Test 2 & 0.0914 & 0.0212 & 0.9088 \\
\cline{2-6}
 & \multirow{2}{*}{Finetuned BERT} & Test 1 & 0.0850 & 0.0194 & 0.9151 \\
 & & Test 2 & 0.0895 & 0.0210 & 0.9107 \\
\hline
\multirow{4}{*}{Team MoraCSE} & \multirow{2}{*}{Rule-based} & Test 1 & 0.6689 & 0.2119 & 0.0177 \\
 & & Test 2 & 0.6809 & 0.2202 & 0.0163 \\
\cline{2-6}
 & \multirow{2}{*}{DL-based} & Test 1 & 0.1983 & 0.0579 & 0.5268 \\
 & & Test 2 & 0.2413 & 0.0789 & 0.4384 \\
\hline
\multirow{4}{*}{Team IndiDataMiner} & \multirow{2}{*}{LLaMa 3.1} & Test 1 & 0.2154 & 0.0881 & 0.5996 \\
 & & Test 2 & 0.2851 & 0.1339 & 0.4879 \\
\cline{2-6}
 & \multirow{2}{*}{Proposed} & Test 1 & 0.1892 & 0.0684 & 0.6288 \\
 & & Test 2 & 0.2640 & 0.1183 & 0.5105 \\
\hline
\multirow{2}{*}{Team NexText} & \multirow{2}{*}{Bi-LSTM} & Test 1 & 0.3450 & 0.0740 &  0.3270 \\
 & & Test 2 &  0.6690 & 0.2270 & 0.0750 \\
\hline
\end{tabular}
\caption{Comparison of Model Performance Metrics Across Teams}
\label{tab:model_comparison}
\end{table*}

The approach of \textbf{Team NexText} \cite{baiju2024romanized} for Malayalam transliteration utilizes Bi-LSTM layers. They trained their model on a substantial combined dataset of 4.3 million transliteration pairs from the Dakshina and Aksharantar datasets. Their data preprocessing strategy operates at the word level, with the model trained on transliteration pairs. During testing, sentences undergo preprocessing, where individual words are extracted, transliterated independently, and then reconstructed into complete sentences. The system carefully preserves non-alphabetic characters, reinserting them after the complete transliteration process. The model architecture is comprehensive and multi-layered. It begins with an encoder input layer capable of processing up to 57 characters in length character sequences. These characters are transformed into 64-dimensional vectors through an embedding layer. A bidirectional LSTM layer then processes this information, capturing sequence patterns from both directions and creating a 256-dimensional representation. This representation is refined through a dense layer, which reduces the dimensionality to a 128-dimensional vector, creating a context vector that feeds into the decoder. The decoder architecture employs a repeat vector layer to duplicate the context vector for each timestep, followed by an LSTM layer for generating hidden states. An attention mechanism enhances the model's ability to focus on relevant parts of the input sequence during the decoding process. The final stage combines the LSTM decoder output with the attention layer through concatenation, feeding into a time-distributed dense layer that produces a probability distribution over possible output characters. The training was conducted on a high-performance computing setup, specifically a single Nvidia DGX A100 GPU with 80 GB RAM. The evaluation was conducted on two distinct test sets: one focusing on general transliteration patterns and another featuring ad-hoc patterns with frequent vowel omissions. While the model demonstrated strong performance on standard typing patterns, it showed some performance degradation when handling ad-hoc typing patterns.

\section{Shared Task Results and Discussion}

The results from the Table \ref{tab:model_comparison} reveal insightful patterns in the performance of various transliteration approaches across Indo-Aryan languages. Team Vectora's Sinhala transliteration models, utilizing Sinhala BERT and Finetuned BERT, exhibited exceptional performance, achieving BLEU scores around 0.91 and remarkably low WER (below 0.09) and CER (around 0.02) for both general and ad-hoc typing patterns. This consistency across diverse test scenarios indicates the robustness of these models in handling both standard transliteration tasks and more challenging cases with omitted vowels. These results position Team Vectora's BERT-based approaches as a strong benchmark for Sinhala transliteration. In stark contrast, Team MoraCSE’s rule-based approach faced significant difficulties, particularly for Sinhala. With a high WER of around 0.67 and very low BLEU scores (below 0.02), it is evident that the rule-based system struggled to address the complexities inherent in Sinhala transliteration. On the other hand, their deep learning (DL)-based approach showed improvement, yielding BLEU scores between 0.43 and 0.52, and WER ranging from 0.19 to 0.24, though these results still fell short of Team Vectora’s performance. For Hindi, Team IndiDataMiner’s models demonstrated moderate success. The proposed model outperformed their LLaMa 3.1 implementation in Test 1 (general typing patterns), achieving a BLEU score of 0.6288 compared to 0.5996. However, both models showed noticeable degradation in performance in Test 2 (ad-hoc typing patterns), reflecting challenges in dealing with irregular vowel combinations. Team NeXText’s Bi-LSTM model for Malayalam exhibited interesting characteristics. While it achieved a relatively low WER of 0.074 for general typing patterns, there was a marked decline in performance for ad-hoc typing patterns (WER increased to 0.227). The high CER values (0.345 and 0.669) and low BLEU scores (0.327 dropping to 0.075) indicated that the model struggled with character-level accuracy, particularly when dealing with irregular typing patterns.

The key observations are listed below.

\begin{itemize}
    \item BERT-based models (Team Vectora) demonstrated superior performance and consistency across different typing patterns, setting a strong benchmark for other transliteration tasks in Indo-Aryan languages.
  \item Deep learning models consistently outperformed rule-based approaches across all languages, highlighting the advantages of neural methods in handling complex language structures.
  \item Ad-hoc typing patterns (Test 2) posed challenges for all models, with performance degradation observed across the board, although the extent varied.
  \item Language-specific characteristics were crucial in model performance, with Sinhala models, particularly Team Vectora’s, proving to be the most robust in handling different typing patterns.
\end{itemize}

These results emphasize that while deep learning models outperform traditional rule-based methods, the specific architecture and training approach, particularly the use of BERT-based models, significantly impacts performance. Team Vectora’s success in Sinhala transliteration provides a promising direction for future work in Indo-Aryan language transliteration tasks.

\section{Conclusion}

In this paper, we present the results of the IndoNLP 2025 shared task, which addresses the challenges of transliteration on Indo-Aryan languages. The participating teams' findings highlight these tasks' ongoing challenges and research gaps. In conclusion, the results of this study underscore the superiority of deep learning approaches, particularly BERT-based models, in handling the complexities of Indo-Aryan language transliteration. Team Vectora's Sinhala models set a new benchmark with their exceptional performance, achieving high BLEU scores and low error rates across both standard and ad-hoc typing patterns. The comparative performance of other teams, such as Team MoraCSE’s deep learning approach and Team IndiDataMiner’s models for Hindi, further confirms the effectiveness of neural-based methods over traditional rule-based systems. While challenges persist in handling irregular typing patterns, the findings highlight the potential of deep learning models to significantly improve transliteration accuracy, particularly when tailored to the specific linguistic characteristics of each language. These results pave the way for more refined and robust transliteration systems, with Team Vectora’s approach offering valuable insights for future work in this domain.

\section*{Ethics Statement}

This paper has benefited from the use of generative AI tools, such as ChatGPT, Notebook LLM and Claude AI, to enhance its clarity and readability. These tools were employed solely for refining language and improving textual coherence without compromising the originality of the research content. The authors take full responsibility for the integrity and accuracy of the content presented in this work.

\bibliography{anthology,custom}
\bibliographystyle{acl_natbib}

\end{document}